\documentclass{article} 
\usepackage{arxiv/collas2025_conference,times}

\usepackage{mathtools}
\usepackage{booktabs} 
\usepackage{easyReview}
\usepackage{microtype}
\usepackage{algpseudocode}
\usepackage{algorithm}
\usepackage{graphicx}
\usepackage{multirow}
\usepackage{booktabs}
\usepackage{xcolor}

\usepackage{hyperref}

\usepackage{amsmath}
\usepackage{amssymb}
\usepackage{mathtools}
\usepackage{amsthm}

\usepackage{subcaption}
\usepackage{wrapfig}
\usepackage{floatflt}

\usepackage{amsmath,amsfonts,bm}

\def\eqref#1{equation~\ref{#1}}

\def\1{\bm{1}}

\DeclareMathAlphabet{\mathsfit}{\encodingdefault}{\sfdefault}{m}{sl}
\SetMathAlphabet{\mathsfit}{bold}{\encodingdefault}{\sfdefault}{bx}{n}

\usepackage{hyperref}
\hypersetup{
    colorlinks=true,
    linkcolor=red,
    filecolor=magenta,
    urlcolor=blue,
    citecolor=purple,
    pdftitle={Overleaf Example},
    pdfpagemode=FullScreen,
    }

\newcommand{\ie}{\textit{i.e.}}

\definecolor{ood_color}{HTML}{0173b2}
\definecolor{al_color}{HTML}{de8f05}
\definecolor{cl_color}{HTML}{029e73}

\title{BOWL: A Deceptively Simple Open World Learner}

\author{
Roshni .R. Kamath${^1}{^,}{^2}$ , Rupert Mitchell${^1}{^,}{^2}$, Subarnaduti Paul${^4}$, Kristian Kersting${^1}{^,}{^2}{^,}{^3}$, Martin Mundt${^4}$ \\
$^1$Department of Computer Science, TU Darmstadt \\
$^2$Hessian Center for AI (hessian.AI)\\
$^3$German Research Center for AI (DFKI) \\
$^4$Department of Mathematics and Computer Science, University of Bremen \\
}

\preprintcopy 

\begin{document}

\maketitle

\begin{abstract}
Traditional machine learning excels on static benchmarks, but the real world is dynamic and seldom as carefully curated as test sets. Practical applications may generally encounter undesired inputs, are required to deal with novel information, and need to ensure operation through their full lifetime - aspects where standard deep models struggle. These three elements may have been researched individually, but their practical conjunction, i.e., open world learning, is much less consolidated. In this paper, we posit that neural networks already contain a powerful catalyst to turn them into open world learners: the batch normalization layer. Leveraging its tracked statistics, we derive effective strategies to detect in- and out-of-distribution samples, select informative data points, and update the model continuously. This, in turn, allows us to demonstrate that existing batch-normalized models can be made more robust, less prone to forgetting over time, and be trained efficiently with less data.
\end{abstract}

\section{Introduction}
Modern deep learning is predominantly developed and assessed on carefully curated datasets, where information is accumulated from observed training inputs, and a dedicated test set is meant to gauge the generalization capabilities. Yet, one needs to only briefly imagine deploying such a pre-trained system into the real world to see that respective ``unseen'' data is unrealistic. Take, for instance, a popular ImageNet model, which may struggle with numerous novel and previously uncaptured objects coming into sight, concepts' appearance differing drastically from the limited training data, or countless variations in conditions surrounding acquisition sensors and environmental factors. While the confined benchmark environments have historically enabled many of the initial algorithmic advances, the next stage of general systems' life cycles needs to account for both the variability and potential informativeness of future experiences in the world - both to guarantee robust deployment and ongoing efficient adaptation. 

To satisfy the requirements of real-world systems, recent works \citep{Boult_owl_2019, mmundt-2020:owll} have postulated at least three essential criteria for the training lifecycle: 
i) the model's ability to statistically differentiate between data points that resemble the known training data and those that are yet unknown; 
ii) the model's ability to actively prioritize the informative data points for future training; 
iii) the model's ability to consolidate information obtained from newly queried data points with past knowledge (see Appendix \ref{app:OWLL_definition} for more details).
Together, a model equipped with these functionalities is able to handle open world environments \citep{chen2018lifelong}.
However, most prior works focus on solving these challenges in isolation from each other, leaving it to the user to integrate them seamlessly. Moreover, there is a lack of solutions to adapt already trained and in-production models for open world learning. 

In this paper, we will show that a batch-normalized deep neural network can already be sufficient to meet the three objectives required for open-world learning. As a ubiquitous and often essential component, batch normalization (BN) \citep{ioffe_batchnorm_covshift} is practically applied before the activation function, taking the pre-activations as the input and standardizing to zero mean and unit variance. In essence, we posit that these statistics of a trained model can serve as a tool to distinguish unseen known from truly unknown data, prioritize acquiring informative data, and minimize catastrophic forgetting induced by continual updates \citep{MCCLOSKEY1989109}. 
To this end, we formulate \textbf{BOWL}: a \textbf{B}atch-norm based \textbf{O}pen \textbf{W}orld \textbf{L}earner. BOWL consists of a common set of 
strategies that cohesively extend any
batch-normalized neural network to open-world learning.
It consists of technically grounded mechanisms constructed using
the running mean and running variance of the batch normalization layer. 
Using the latter as a simple Gaussian distribution, BOWL enables robust prediction through a measure of statistical deviation to flag unseen, unknown data. To query novel data, it leverages batch-norm statistics to compute information density. Finally, to enable continual updates, it adaptively manages a dynamic memory.

In summary, our contributions are as follows:
\begin{itemize}
\item By leveraging the ubiquitous use of batch normalization, we introduce the first monolithic and easy-to-implement framework to extend existing models to open-world learning: BOWL.

\item We show how the statistics of the batch normalization layer serves a cohesive basis to equip models with the capabilities of out-of-distribution detection, active learning, and continual learning.

\item Through a sequence of experiments, we show that BOWL is suitable for open-world learning. We empirically corroborate that i) its dynamic memory management allows to learn efficiently with less data and entails less optimization steps, ii) its three batch-norm based components are essential contributors to enable open world learning, iii) its cohesive design leads to robustness under adverse data conditions and open world settings. 
\end{itemize}

\section{Related Works}
This section discusses approaches put
forth to alleviate the problems of the closed world setting independently --- respectively:
out-of-distribution detection, active learning, and continual learning.
We also provide a preview of the design choices for BOWL and how it
can be differentiated.  

\textbf{Out of Distribution Detection}: Out of Distribution (OoD) Detection \citep{oodsurvey_yang2021, mukhoti2021deterministic} is used to reject samples that statistically deviate from the training distribution.
Open Set Recognition (OSR) \citep{Boult_2013_osr, osr_survey_geng_2021} emerged as a specific idea to minimize the volume of unknown samples (unknown unknowns) outside the domain of known samples (known unknowns) by creating a "boundary" with the help of a measurable recognition function. Building on the idea of OSR, \citet{Bendale_2016_CVPR, Mundt_2019_ICCV} and \citet{Joseph_2021_CVPR} designed algorithms to differentiate between known and unknown data instances. OoD is not generally an open-set classifier but can nevertheless be incorporated as a detector to differentiate learnable labels proportionately.  
Such methods are helpful in open-world applications, where a model 
comes across corrupted and redundant data, minimally labeled data, data distribution shifts, and fluid task boundaries. 
In our framework, we design a simple, yet effective, outlier hypothesis based on readily available batch norm statistics to reject samples that largely stray from the learned distribution. This hypothesis also enables the identification of new classes without additional operations and lets an oracle provide labels.

\textbf{Active Learning}: Active learning \citep{cohn_ghahramani_jordan_1996} 
allows a model to query for 
novel data points for inclusion into future training. 
This process is beneficial in scenarios where labeled data is limited, expensive, or not fully initially available.
The most informative points 
are determined using the ``acquisition function'', where 
data points that can improve the 
performance (on potentially unseen datasets) are desirable.
This, in turn, aims to improve generalization
and to reduce the uncertainty of the model \citep{guo:active_learning}. 
Predominantly, the model is then trained from scratch by interleaving newly acquired data points with the old instances, hence assuming the availability of the old dataset. 
One can remove the latter dependency 
on the availability of old data 
by synergizing the active learning strategies
with continual learning, as done in e.g. \citet{ayub2022fewshot}. 
BOWL derives its \textit{acquisition function} from batch-norm statistics, drawing inspiration from earlier works on information density \citep{burr2008_informatn_density}.

\textbf{Continual Learning}: 
A model undergoes catastrophic forgetting \citep{MCCLOSKEY1989109} when  
representations are formed from 
only new data, in turn
leading to a performance drop
on previous training data.
Multiple research works on continual learning (CL) \citep{chen2018lifelong} 
have engineered three pathways to mitigate this phenomenon, namely: 
i) regularization --- \ie{} tracking parameter importance, 
ii) rehearsal --- \ie{} replaying past data, and 
iii) architecture-based methods --- \ie{} retaining task-relevant regions of a neural network. Respectively, 
\citet{ewc,vcl,synaptic_intel} and \citet{benjamin2019measuring} preserve learned representations
by regularizing the loss, such that the new parameters remain close to the previous ones.
Replay based methods, such as \citet{ER_2019,icarl}, store and use old data while training on new tasks.
Architectural methods \citep{pnns,mntdp} 
reflect on how old parts of the network can be effectively used to solve new tasks.
\citet{survey_farquhar2019robust, lange_continual_learning_survey, survey_wang2023comprehensive} 
further provide a detailed summary of different methods adopted throughout continual learning. 
Owing to the simplicity and efficacy of replay-based methods \citep{ER_2019,deepgenerative_replay}, 
BOWL maintains a 
memory buffer. 
However, unlike other methods that balance data in the memory, BOWL dynamically measures the 
usefulness of data based on the batch norm statistics and composes its memory adaptively.

\section{Batch-Normalized Models for Open World Learning}
\label{sec:bowl_modules}

We show that a neural network with arbitrary numbers of batch normalization layers is an open-world learners
We formulate the OoD, Active learning, and CL strategies using the batch statistics available in the BN layer.
We then chain them with each other to produce a monolithic open-world learner.
The ensuing subsections, \S \ref{sec:ood_module}, \ref{sec:active_query_module} and \ref{sec:continual_train_module}, provide the mathematical details of the respective strategies. 
We use the three strategies to integrate and outline the BOWL framework for open-world learning in \S \ref{sec:BOWL_glance}

\subsection{Preliminaries: Batch Normalization Layers and Gaussian Uncertainty}
\label{sec:batch_norm_gaussian}

\begin{figure*}[t]
\centering
\includegraphics[width=0.9\columnwidth]{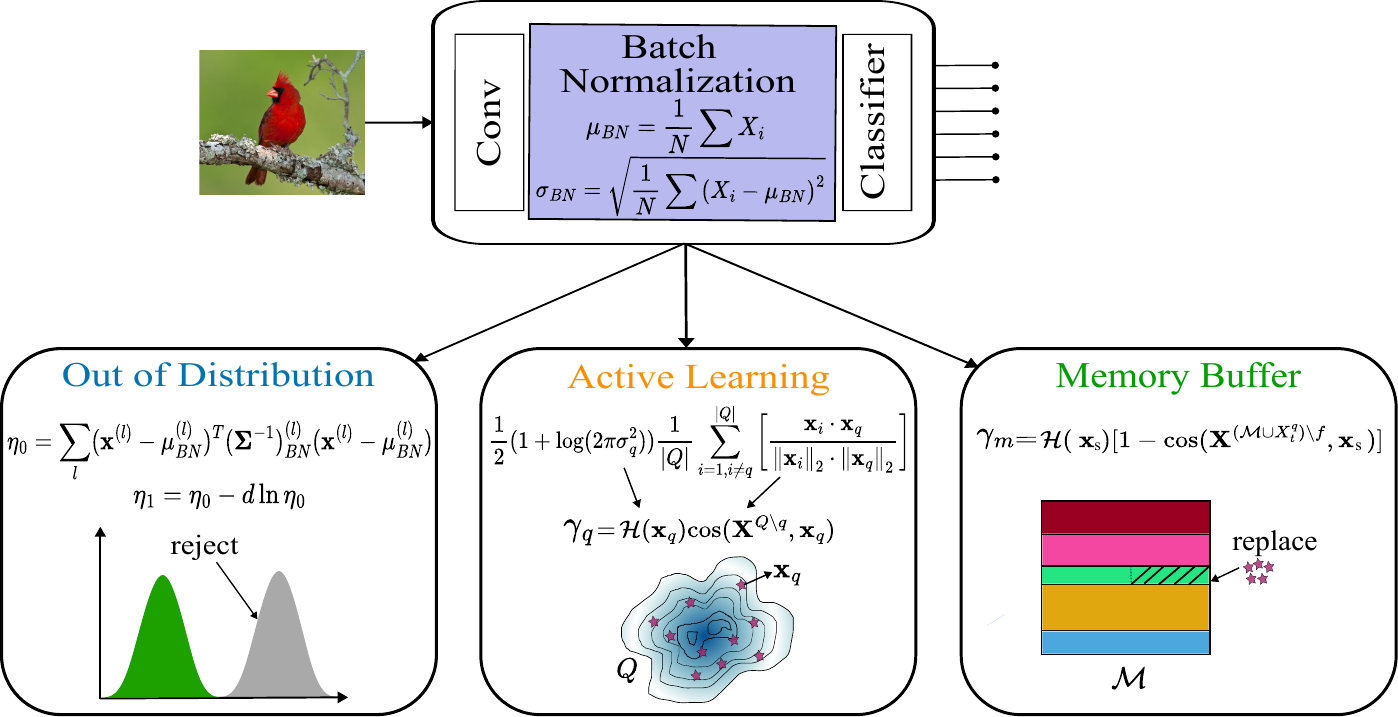}
\caption{\textbf{Visualization of individual BOWL components, all rooted in a common frame of batch-normalization.} For adapting a deep neural network to open-world learning, we use the running mean ($\mu_{BN}$) and the running average ($\sigma_{BN}$) from the batch normalization layer. The Out of Distribution module assigns a score $\eta_1$ to the data points, used for rejecting outliers. The Active Learning module chooses top $\gamma_q$ scoring data points from the available pool $Q$ to include into further training. The Continual Learning module stores task-suitable points based on $\gamma_m$ in the memory buffer $\mathcal{M}$.} 
\label{fig:bowl_illustration}
\end{figure*}

The batch normalization layer \citep{ioffe_batchnorm_covshift} uses an estimate of the mean $\mu_x$ and standard deviation $\sigma_x$
of its input $x$ to produce a shifted and scaled output $y$:
\begin{equation}
    y = \gamma \frac{(x - \mu_x)}{\sqrt{\sigma^2_x + \epsilon}} + \beta
\end{equation}
where $\epsilon$ is some small constant,
and $\gamma$ and $\beta$ are learnable parameters initialized to $1$ and $0$ respectively. It is an indispensable component in the majority of neural networks that improves learning and generalization by hypothesized internal covariate shift reduction \citep{ioffe_batchnorm_covshift}.

The key property of the batch norm layer from our perspective is that,
for every point in our architecture with a batch norm layer,
we automatically know the mean and standard deviation of the intermediate values at that point.
This allows us to model these intermediate values $x$ as being distributed according to
a Gaussian distribution with mean $\mu_x$ and variance $\sigma_x^2$. 
We would like to stress that no simplifying assumptions are made on the use of batch normalization in our work,
and that the results derived are applicable to any network
architecture that makes use of batch normalization exactly as it appears
in practical applications.
Although there may exist disagreement on the precise contribution of BN to optimization \citep{bn_santurkar_NEURIPS2018, bn_rank_collapse_NEURIPS2020, bn_cov_shift_NEURIPS20201, bn_cov_shift_2020},
the usefulness of the statistics derived from the BN layer remains unaffected for open world learning.

\subsection{Leveraging Batch-norm Statistics for Out of Distribution Detection}
\label{sec:ood_module}
The \textit{OoD} module is designed to reject highly unusual data-points before they are placed into the queryable pool.
This is useful because some of the data may be corrupted or may simply be irrelevant.
We do not want to waste the capacity of our continual learner on trying to fit these data-points.\\
To determine if an input is unusual we can compare the intermediate values $\mathbf{x}^{(l)}$
at every batch norm layer $l$ with their expected distribution $\mathcal{N}^{(l)}_{BN}$
according to the batch norm layer. The simplest sense in which the values $\mathbf{x}^{(l)}$ could be considered unusual
is if they are low probability according to $\mathcal{N}_{BN}^{(l)}$.
We could assign the activations a score $\eta_0$ given by
\begin{equation}
    \eta_0 = \sum_l \left(\mathbf{x}^{(l)} - \mathbf{\mu}_{BN}^{(l)}\right)^T \left(\mathbf{\Sigma}_{BN}^{-1}\right)^{(l)} \left(\mathbf{x}^{(l)} - \mathbf{\mu}_{BN}^{(l)}\right)
\label{eq:eta_0}
\end{equation}
which is proportional to the negative log probability of $\mathbf{x}$ according to $\mathcal{N}_{BN}$ plus a constant.
This is effectively a measure of whether or not the values $\mathbf{x}^{(l)}$ are larger in magnitude than expected.
This score is well motivated as a way of discarding outliers,
and is effective at discarding outliers which produce very large intermediate values.
Unfortunately, some data items we would like to discard produce intermediate values which are very small instead of very large.
In order to address both cases simultaneously,
we use a simple modification $\eta_1$ of $\eta_0$ given by
\begin{equation}
    \eta_1 = \eta_0 - d \ln{\eta_0}
    \label{eq:eta_1}
\end{equation}
where d is the dimension of $\mathbf{x}$.
It can be seen that $\eta_1$ is large for both unusually large and unusually small intermediate values.
We compute $\eta_1$ for each batch in the dataset and allow batches into the queryable pool if their corresponding value of $\eta_1$ is below some threshold $\tau$.
In the Appendix \ref{app:ood_derivation} we show how both $\eta_0$ and $\eta_1$ can be derived
from the application of Bayes' theorem,
but under different assumptions about the variance of intermediate values produced by outlying data.
\begin{wrapfigure}[21]{r}{0.55\textwidth}
    \begin{minipage}{0.49\linewidth}
    \includegraphics[width=\linewidth]{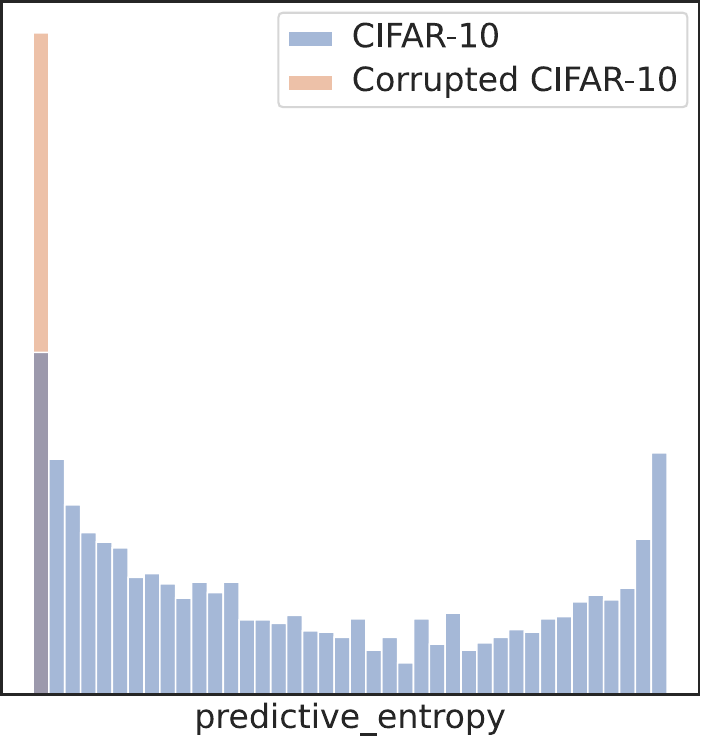}
    \end{minipage}%
    \hfill
    \begin{minipage}{0.49\linewidth}
    \includegraphics[width=\linewidth]{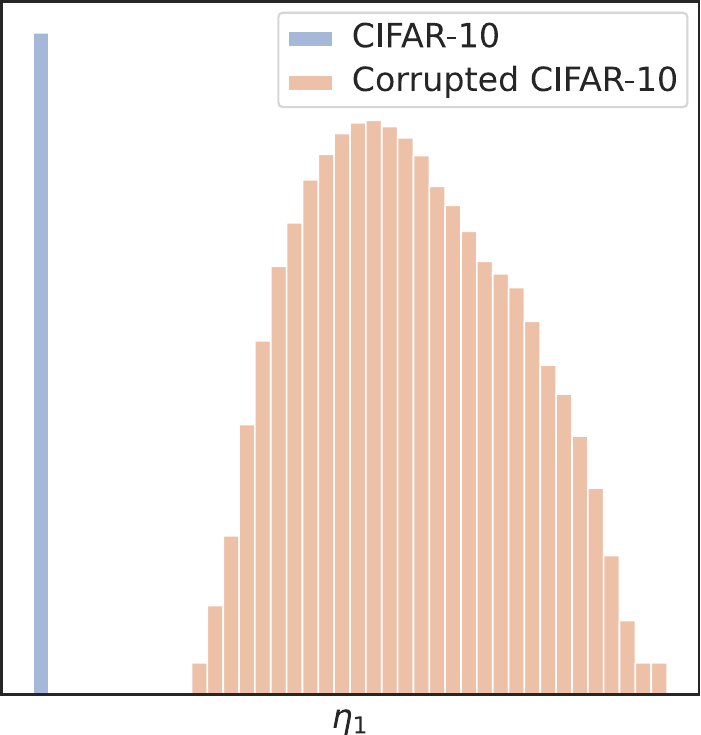}
    \end{minipage}
\caption{\textbf{Batch-norm based OOD detection is able to identify corrupted examples}. A Resnet-18 model trained on Split CIFAR-10. Blue corresponds to the score on IN-distribution (CIFAR-10) and yellow corresponds to the score on OUT-of-distribution (CIFAR-10 corrupted with various forms of perturbations). The difference between the OoD and IN-distribution histograms is pronounced in the case of BOWL.}
\label{fig:ood_results}
\end{wrapfigure}
Specifically, $\eta_0$ corresponds to assuming that this variance is large with respect to that of the inlying data,
whereas $\eta_1$ corresponds to assuming that it is simply unknown.
We also briefly discuss on how we set a threshold $\tau$ that allows data into the queryable pool.

To provide intuition behind the OoD detection with equation \ref{eq:eta_1}, we validate its ability to distinguish between in-distribution (ID) images from CIFAR-10 and OoD images from Noisy CIFAR-10.
As can be seen in Figure \ref{fig:ood_results}, the use of equation \ref{eq:eta_1} fully separates ID images (visible as a single bar in the right panel) from OOD images in the case of Noisy CIFAR-10-C.
In comparison, we show the same histograms for the known alternative method of predictive entropy (PE).
It can be seen in the left panel that PE produces inverted (worse than random) results for Noisy CIFAR-10, as the model is overconfident on corrupted images.
In summary, the OoD detection abilities of BOWL are sufficient and decisively superior to the obvious alternative due to the ability to examine the distribution of intermediate values from the network instead of just outputs.


\subsection{Leveraging Batch-Norm Statistics to Actively Query Informative Samples}
\label{sec:active_query_module}
The purpose of the query function is to select data from the pool to be labeled and used in training the model.
We would like to select data whose labels will be most informative about the task(s) the model is being optimized to solve.
There are two major reasons it might be less worthwhile to label a particular datum.
Firstly, the datum might be too similar to other data for which we already have labels.
Secondly, it might be unusual enough that it is mostly unrelated, not only to the data we already have labeled but also to any data we expect to need to predict labels for in the future (e.g., at test time).
We attempt to capture these two failure modes with two heuristics.

For the first heuristic $\alpha_q$, we use the following quantity:
\begin{equation}
    \begin{split}
    \alpha_q
    &= \frac{1}{2}(1+\log (2 \pi \sigma^2_{q}))
    \end{split}
\end{equation}
Here, $\sigma_q$ is the standard deviation across pixels, channels, and layers of the post-batch-norm intermediate activations of the network when given the input $\mathbf{x}_q$, i.e.
\begin{equation}
    \sigma_q^2 = \frac{1}{|L|}\sum_{l \in L} \frac{1}{|C_l|} \sum_{c \in C_l} \frac{1}{|P_l|} \sum_{p \in P_l} a_{lcpq}^2
\end{equation}
where $L$ is the set of layers and $C_l$ and $P_l$ are the sets of channels and pixels for layer $l$, and $a_{lcpq}$ is the post-batch-norm activation at pixel $p$, channel $c$ and layer $l$ when the input to the network is $\mathbf{x}_q$.
This is a measure of how spread out the intermediate activations are relative to their distribution on the existing labeled data, since the batch norm layer normalizes its inputs with respect to the statistics of that data.
We use this measure of dissimilarity in intermediate activation space as a proxy for dissimilarity in terms of what information their labels would give us about the task.

\begin{figure*}[t]
\includegraphics[width=\linewidth]
{images/bowll_ood_cifar10_vs_imagenet_log}
\caption{\textbf{BOWL adaptively selects informative examples to populate a diverse memory buffer.} Left: Amount of queried samples (y-axis) across tasks ($T_1$ to $T_4$ labels) over time (batch updates, x-axis) on Split CIFAR-10 with ResNet-18, where the green bars represent the number of samples queried for the new task, and the purple curve shows the test accuracy. Right: Memory buffer composition across the tasks. GDUMB's artificial balance yields $52\%$ accuracy whereas BOWL's knowledge-driven adaptive strategy achieves $63\%$ accuracy.}
    \label{fig:memory_buffer_stats}
\end{figure*}

For the second heuristic $\beta_q$ we use the following normalized inner product in input space:
\begin{equation}
    \begin{split}
    \beta_q
    &=
    \frac{1}{|Q|}\sum_{i = 1, i \neq q}^{|Q|}\left[\frac{\mathbf{x}_i \cdot \mathbf{x}_q}{\left\|\mathbf{x}_{i}\right\|_2 \cdot\left\|\mathbf{x}_q\right\|_2}\right]
    \end{split}
\end{equation}
where $Q$ is the queryable pool and $\mathbf{x}_q$ is the datum we are evaluating.
This quantity captures a notion of similarity averaged over the queryable pool.
It can be seen to reach a value of $1$ for the case where $\mathbf{x}_q$ is identical to every other $\mathbf{x}_i$, and a value of $0$ for the case where $\mathbf{x}_q$ is orthogonal to them.
Taking the pool to be a proxy for the kinds of data we want the model to perform well on addresses the second failure mode, \ie avoiding too unusual points.

To combine these two heuristics, we multiply them, giving the query score formula:
\begin{equation}
    \gamma_q
    = \underbrace{\frac{1}{2}(1+\log (2 \pi \sigma^2_{q}))}_{\mathcal{H}(\mathbf{x}_q)}* 
    \underbrace{\frac{1}{|Q|}\sum_{i = 1, i \neq q}^{|Q|}\left[\frac{\mathbf{x}_i \cdot \mathbf{x}_q}{\left\|\mathbf{x}_{i}\right\|_2 \cdot\left\|\mathbf{x}_q\right\|_2}\right]}_{{\cos}({\mathbf{X}}^{Q\backslash q}, \mathbf{x}_q)}
    \label{eq:active_query}
\end{equation}
where $\gamma_q=\alpha_q \beta_q $ is the query score for datum $\mathbf{x}_q$. In this way, the datum's novelty assessed through the entropy term ($\mathcal{H}$) is weighed by its typicality. 
The choice of multiplying the heuristics does not immediately require us to assign a weight to the two heuristics,
and we did not find it necessary to look for a way to introduce any such weighting.

\subsection{Leveraging Batch-Norm Statistics to Alleviate Catastrophic Forgetting} 
\label{sec:continual_train_module}

When a new stream of data is encountered,
we want the model to learn on this stream gradually 
such that the data points are close to the data available in the memory. 
For the model to retain already learned representations while continuing to adapt, 
we must adopt strategies that combat catastrophic interference while training on new data.
To achieve this, 
we draw inspiration from our active learning mechanism to dynamically manage a memory buffer. 
Firstly, we identify notable samples from the queried data-points using the selection function
discussed in section \ref{sec:active_query_module},
as candidates to replace existing data-points in the memory buffer. 
Following the query's spirit in Eq. \ref{eq:active_query}, we then sample high-scoring points of size $|\mathcal{M}|$ and 
assign a memory score $\gamma_m$:
\begin{equation}
    \gamma_m = \mathcal{H}(\mathbf{x}_s)*[1- {\cos}({\mathbf{X}}^{(\mathcal{M} \cup X_t^q)\backslash f}, \mathbf{x}_s)].
\label{eq:memory_buffer}
\end{equation}

With this score, BOWL's fixed size memory buffer is dynamically repopulated by integrating newly acquired samples with existing ones. by retaining the top $|\mathcal{M}|$ points based on their $\gamma_m$ score. After replacing old points with new data, we then train the model on the memory buffer for 1 epoch, \ie one pass through the buffer.
Importantly, we note that BOWL only ever trains the model on the examples in the memory buffer. To dynamically reevaluate a datum's value in the memory, we additionally retain the entropy of the stored data points.
Over time, older and no longer necessary samples will thus be replaced with novel information, but due to balancing of equation \ref{eq:memory_buffer}, data points that remain critical to the model will also prevail. 
We hypothesize that such gradual incremental updates can minimize catastrophic interference 
rather than abruptly introducing the data that has not yet been seen.

We now highlight the efficacy behind BOWL's \textit{active query} ability and its \textit{dynamic memory buffer}
on Split CIFAR-10 with ResNet-18 for a classification task.
At each training step of Figure \ref{fig:memory_buffer_stats}, we acquire new samples from an unlabeled pool $\mathcal{Q}$ of size $\mathcal{B}=256$ to dynamically populate a memory buffer $\mathcal{M}=5000$.
In the left subplot, 
we plot the exponential moving average
of the samples procured by the AL strategy (green bars)
across the 4 tasks ( $T_1$ through $T_4$).
For each task, 
the model populates the memory buffer with these samples, while the test accuracy (purple line) shows the respective learning and forgetting cycles.
Each task begins with a rapid performance gain, with accuracy improving from approximately 30\% to 60-75\% within the first 10-15 batch updates. 
This is followed by a controlled adaptation phase where 
BOWL demonstrates its ability to maintain performance while incorporating new information. We can see that while BOWL could potentially query up to 256 new examples, it progressively requires less instances as more knowledge is build up, highlighting the value of its dynamic memory management. 
The right subplot shows the respective class composition of the memory buffer through stacked bar plots 
for GDUMB (G) and BOWL (B).
Whereas both methods retain core categories of the CIFAR-10 dataset, BOWL adapts the ratios while preserving critical examples. 
Overall, BOWL's memory buffer undergoes gradual adjustments that lead to stability in training accuracy. Its controlled sample selection and replacement strategies ensure that classes that require more critical examples in memory, are allocated more space in the buffer. This is fundamentally different from the majority of replay mechanisms, that assume all classes and data points are equally valuable. 
Due to the balance of new learning and memory consolidation, BOWL therefore achieves competitive performance that outshines GDUMB.


\subsection{The BOWL algorithm}
\label{sec:BOWL_glance}

\begin{floatingfigure}[l]{0.55\columnwidth}
\vspace{-0.65cm}
\begin{minipage}{0.55\columnwidth}
\begin{algorithm}[H]
    \caption{BOWL Framework}
    \label{alg:bowl}
    \begin{flushleft}
    \textbf{Input}: Trained model $f_{\theta_{t-1}}$, data $\{x_t^i, y_t^i\}_{i=1}^{N}$ for $t=1, 2, \ldots, T$, acquisition batch size $\mathcal{B}$ \\
    \textbf{Initialize}: Memory Buffer $\mathcal{M} \leftarrow \{x_0^i, x_0^i\}_{i=1}^{|\mathcal{M}|}$
    \end{flushleft}
    \begin{algorithmic}[1] 
        \For{each timestep $t$ with data $\{x_t^i, y_t^i\}_{i=1}^{N}$}
            \State Identify in-distribution data points using $\eta_1$ from eq. \ref{eq:eta_1}: 
            \Statex \hspace{2em} $\{x_t^k, y_t^k\} = \textcolor{ood_color}{\text{OoD Detection}}(f_{\theta_{t-1}}, x_t)$ 
            \State Populate the candidate pool $\mathcal{Q} \leftarrow \{x_t^k, y_t^k\}$
            \While{$\mathcal{Q}$ is not empty}
                \State Select top scoring informative samples using eq. \ref{eq:active_query}: 
                \Statex \hspace{2em} $\{x_t^q, y_t^q\} \leftarrow \textcolor{al_color}{\text{Active Query}}(\mathcal{Q}, \mathcal{B})$
                \State Update the \textcolor{cl_color}{Memory Buffer} using eq. \ref{eq:memory_buffer}:
                \Statex \hspace{2em} $\mathcal{M} \leftarrow (\mathcal{M} \setminus \text{replace samples}) \cup \{x_t^s, y_t^s\}$, 
                \State Train the model $f_{\theta_{t}}$ on $\mathcal{M}$ for 1 epoch.
            \EndWhile 
        \EndFor
    \end{algorithmic}
\end{algorithm}
\end{minipage}
\end{floatingfigure}

For completeness, we lay out the entire BOWL framework in Algorithm \ref{alg:bowl}, which also depicts the organization of every module. 
We begin with a previously trained model $f_{\theta_{t-1}}$ that contains batch norm layers.
At each timestep, BOWL processes the incoming stream 
through three mechanisms : i) First, it employs the out-of-distribution module (\ref{sec:ood_module})
to identify and filters instances that align with the model's current knowledge distribution, using the threshold 
$\eta_1$,
ii) Second, from these filtered samples, it actively selects the most informative instances based on the score $\gamma_q$ (\ref{sec:active_query_module}), later processing them in batches,
iii) As the last step, these selected samples are incorporated into a fixed-size memory buffer denoted by $\mathcal{M}$. During this integration, older or less relevant samples are replaced based on $\gamma_m$ scores(\ref{sec:continual_train_module}). Once the buffer is updated, the model is subsequently trained for one epoch on the revised memory. 

\bigskip

\section{Experiments}
In this section, we demonstrate BOWL's suitability as an open world learner and corroborate its design choices. Most importantly, we provide empirical support to highlight that BOWL's: \\
1) adaptivity leads to competitive accuracy while consuming less data and requiring less overall optimization steps, \\
2) triplet of batch-norm based components each contributes meaningfully to overall performance, \\
3) cohesive design leads to robustness under adverse data conditions and open world settings. \\
Precise details on 
the experimental setups, task splits and 
comparison methods are deferred
to the Appendix \ref{app:training_configs}.

\subsection{Optimization efficiency}
\label{sec:improved_batch_updates}

\begin{figure}[t]
    \centering
        \includegraphics[width=0.95\columnwidth]{images/bowl_cifar10_optimization_steps_120_longer.pdf}
        \captionof{figure}{\textbf{BOWL adapts rapidly by selecting essential samples and requires substantially less optimization steps.} 
        BOWL yields competitive accuracy ($\approx60\%$ y-axis) while using only 25\% of training data
        and 12\% of the optimization steps (x-axis) compared to baseline methods.
        Despite sharing the same memory buffer size (5000 samples) with other replay based contenders, BOWL demonstrates faster adaptation and better performance across all tasks ($T_1$ to $T_4$).}
        \label{fig:bowl_cifar10_optimization_steps}
\end{figure}

A key advantage of BOWL lies in selectively prioritizing data points that are not only
exemplars from the current task 
but also in congruence with the past training data. This leads to less required data and reduces the number of needed optimization steps.
To substantiate this claim, we present the results of training a ResNet-18 model on Split CIFAR-10 in the Figure \ref{fig:bowl_cifar10_optimization_steps}.
The x-axis describes the number of optimization steps divided into segments of tasks $T_1, T_2, T_3$, and $T_4$ against the average test accuracy on the y-axis.
We plot the test accuracies obtained over average of 100 steps
and map to their corresponding x-axis positions.

In the initial training phase for task $T_1$, BOWL attains $\approx85\%$ accuracy with fast convergence of $\approx750$ optimization steps. 
This pattern is observed across all tasks
with BOWL achieving high task accuracies rapidly. 
However, methods like ER \citep{ER_2019}, CLSER \citep{arani_complementarylearning} and Kprior \citep{kpriors_khan2021} train on the entire new data as well as 
the memory buffer (5000 samples)
for 120 epochs, 
hence requiring 21094 steps for all the tasks in total.
Despite consuming such high resources, the final accuracy is lower than that of BOWL, which consumes much less data and $\approx$ 2700 overall optimization steps.
When encountering new tasks, 
BOWL's mechanisms can assess data eligibility and identify potential novel tasks allowing the model 
to adjust its learning accordingly.  
As outlined in \S\ref{sec:active_query_module} and \S\ref{sec:continual_train_module}, the benefits are twofold here:
i) the model is able to periodically refine 
its understanding of what constitutes informative samples,
ii) the periodic updates ensure that relevant samples remain in the buffer as the model's knowledge increases. 
This iterative process helps BOWL make increasingly better selection decisions as the model's feature representations improve through 
training
For CIFAR-10, BOWL demonstrates superior performance, 63\% vs 51\% with GDUMB \citep{gdumb},
even when similarly exclusively trained only on the memory buffer. It even beats ER, which first trains on all data and then retains a memory buffer, despite using fewer data and fewer gradient updates.
In contrast to BOWL, the other methods CLSER, Kprior, GDUMB, and ER rely primarily on rehearsal mechanisms that 
populate the respective buffer once per task
to prevent catastrophic forgetting. 
While these balanced approaches allow the model to retain past knowledge, 
they cannot identify which samples remain informative as learning progresses, and thus use computational resources inefficiently. 
This limitation becomes increasingly apparent as tasks progress from $T_1$ to $T_4$, where varied rehearsal-based methods struggle to maintain performance.
BOWL’s strong performance across tasks, especially in the initial phases of learning a new task, suggests that it’s making efficient use of the data it receives. The rapid increase in test accuracy at the start of new tasks (such as in $T_2$ and $T_3$) highlights that active learning is essential in BOWL to learn new tasks faster than methods with random sampling (like GDUMB or ER).

Table \ref{tab:bowl_dataset_stats} presents the final test accuracy of BOWL on Split CIFAR-10 and Split ImageNet-200 datasets trained with Resnet-18 \citep{resnet_2016} and EfficientNet-B2 \citep{tan2019efficientnet} respectively.
The table also reports the number of observed data points (\# ODP ($\downarrow$))
and optimization steps (Steps ($\downarrow$)) 
undertaken by each method.
After the final step in $T_4$, BOWL reaches approximately 63\% accuracy, beating all other methods on Split CIFAR-10.
For Split ImageNet-200,
BOWL's 13.90\% accuracy outshines ER (11.93\%) and Kprior (12.73\%), while using only $\approx30\%$ of the samples. On both datasets, BOWL cuts down the number of optimization steps from approx. 21000 and 246000 optimization steps, to merely 2696 and 26641 on split CIFAR-10 and Imagenet-200 respectively. This accounts for $\approx10\%$ of gradient updates.
It is critical to note that both BOWL and GDUMB train on
fixed-size (5000) memory buffer, however GDUMB under-performs majorly. 
Datasets like ImageNet-200 have varying 
complexity levels with diverse class distribution. 
GDUMB's rigid approach 
to maintain fixed class bins
with a fixed-size memory buffer
is  insufficient.
BOWL also learns only from the memory buffer,
however it
iteratively queries new samples according to informativeness and dynamically populates the buffer
for the model to train on. 
This also means that despite the fixed size of the memory buffer (5000),  
it can strategically introduce more samples into the memory buffer and later replace them
as demanded by the complexity of the dataset.
This improves the performance (13.90\%) while maintaining 
the same constant memory size during training.
Beyond this observed training efficiency under strong memory constraints, the synergy of BOWL's three components will also become even more apparent in the subsequent experimental subsections.

\begin{table*}[t]
\centering
     \captionof{table}{
        \textbf{BOWL's periodic memory updates enable efficient learning under memory buffer constraints.} Comparison of BOWL with other continual learning methods across Split CIFAR-10 and Split ImageNet-200. BOWL outperforms all the methods on CIFAR-10, while training with significantly less samples (\#ODP: number of observed data points) and requiring substantially less optimization steps (Steps). On ImageNet, it achieves twice the performance of GDUMB, by striking a better trade-off. Best values are in bold, second-best are underlined.}
        \label{tab:bowl_dataset_stats}
        \resizebox{\columnwidth}{!}{
        \begin{tabular}{lccccccccc}
        \toprule
        \multirow{2}{*}{Method} & \multicolumn{4}{c}{Split CIFAR-10} & \multicolumn{4}{c}{Split ImageNet-200} \\
        \cline{2-9} 
         & Accuracy $(\uparrow)$ & Steps $(\downarrow)$ & \#ODP $(\downarrow)$ & Buffer Size & Accuracy $(\uparrow)$ & Steps $(\downarrow)$ & \#ODP $(\downarrow)$ & Buffer Size \\
        \hline
        Finetune & 19.80\tiny{$\pm0.00$} & 18750 & 40000 & - & 6.50\tiny{$\pm0.00$} & 234375 & 100000 & -  \\
        EWC & 19.72\tiny{$\pm0.00$} & 18750 & 40000 & - &  8.39\tiny{$\pm0.00$} & 234375 & 100000 & -  \\
        ER & \underline{60.37\tiny{$\pm0.02$}} & 21094 & 40000 & 5000 & 11.93\tiny{$\pm0.00$} & 246093 & 100000 & 5000 \\
        GDUMB & 51.96\tiny{$\pm0.00$} & \underline{9375} & \textbf{4688} & 5000 & 6.88\tiny{$\pm0.00$} & \underline{117188} & \textbf{4688} & 5000 \\
        CLSER & 19.00\tiny{$\pm0.02$} & 21094 & 40000 & 5000 & 5.51\tiny{$\pm0.08$} & 246093 & 100000 & 5000 \\
        Kprior & 28.14\tiny{$\pm0.03$} & 21094 & 40000 & 5000 & \underline{12.73\tiny{$\pm0.15$}} & 246093 & 100000 & 5000 \\
        BOWL & \textbf{63.74\tiny{$\pm1.79$}} & \textbf{2696\tiny{$\pm205$}} & \underline{10207\tiny{$\pm570$}} & 5000 & \textbf{13.90\tiny{$\pm0.02$}} & \textbf{26641\tiny{$\pm17$}} & \underline{24085\tiny{$\pm2496$}} & 5000 \\
        \bottomrule
        \end{tabular}
        }
\end{table*}

\subsection{Ablation Analysis on Split CIFAR-10}
\label{sec:ablation_study}
\begin{figure*}[t]
    \begin{minipage}{0.5\columnwidth}
\setlength{\tabcolsep}{1pt}
\captionof{table}{\textbf{Ablation analysis of BOWL on Split CIFAR-10}. We report the test accuracy at the end of every timestep. ``$\times$'' indicates that the module is omitted and ``$\checkmark$'' indicates otherwise. \textbf{iv} represents the complete BOWL framework with all its modules present. The final performance highlights that the inclusion of each individual component is meaningful to the overall performance. This suggests that effective learning in open-world scenarios requires a wholistic approach that addresses identification of relevant samples, gauging data informativeness, and adaptive memory management. Notably, that even if the absence of OoD module initially appears to have on-par accuracy, 
the complete framework is required to achieve robust performance across all evaluated scenarios --- as is highlighted in Figure \ref{fig:bowl_open_world_learning}.}
\centering
\resizebox{\linewidth}{!}{
\begin{tabular}{@{}ccccccccc@{}}
\toprule
\multirow{2}{*}{} &
\multicolumn{3}{c}{Modules} &
  \multicolumn{4}{c}{Accuracy} & 
  \multicolumn{1}{c}{\# Samples} \\ \cmidrule(lr){2-4} \cmidrule(lr){5-8}
   & OoD & AL & CL &
  \multicolumn{1}{c}{t=1} &
  \multicolumn{1}{c}{t=2} &
  \multicolumn{1}{c}{t=3} &
  \multicolumn{1}{c}{t=4} &
  \\ 
  \midrule
i & $\times$ & $\checkmark$ & $\checkmark$ & 
\multicolumn{1}{c}{84.51\tiny{$\pm$0.91}} & \multicolumn{1}{c}{81.32\tiny{$\pm$0.37}} & \multicolumn{1}{c}{72.46\tiny{$\pm$0.52}} & \multicolumn{1}{c}{66.14\tiny{$\pm$0.57}} & \multicolumn{1}{c}{10581\tiny{$\pm18$}} \\ 

ii & $\checkmark$ & $\times$ & $\checkmark$ & 
\multicolumn{1}{c}{84.27\tiny{$\pm$0.56}} & \multicolumn{1}{c}{58.50\tiny{$\pm$0.49}} & \multicolumn{1}{c}{40.11\tiny{$\pm$0.10}} & \multicolumn{1}{c}{33.68\tiny{$\pm$0.39}} & \multicolumn{1}{c}{4693\tiny{$\pm0$}} \\ 

iii & $\checkmark$ & $\checkmark$ & $\times$ & 
\multicolumn{1}{c}{41.06\tiny{$\pm$0.78}} & \multicolumn{1}{c}{39.32\tiny{$\pm$7.70}} & \multicolumn{1}{c}{24.41\tiny{$\pm$9.57}} & \multicolumn{1}{c}{15.32\tiny{$\pm$4.44}} & \multicolumn{1}{c}{39641\tiny{$\pm317$}} \\ \midrule

iv & $\checkmark$ & $\checkmark$ & $\checkmark$ & 
\multicolumn{1}{c}{84.21\tiny{$\pm$0.74}} & \multicolumn{1}{c}{79.77\tiny{$\pm$1.84}} & \multicolumn{1}{c}{69.32\tiny{$\pm$0.81}} & \multicolumn{1}{c}{63.74\tiny{$\pm$1.79}} & \multicolumn{1}{c}{10207\tiny{$\pm520$}} \\
\bottomrule
\end{tabular}
}
\label{tab:ablation_study_1}
    \end{minipage}
    \hfill
    \begin{minipage}{0.47\columnwidth}
    \centering
    \includegraphics[width=\linewidth]{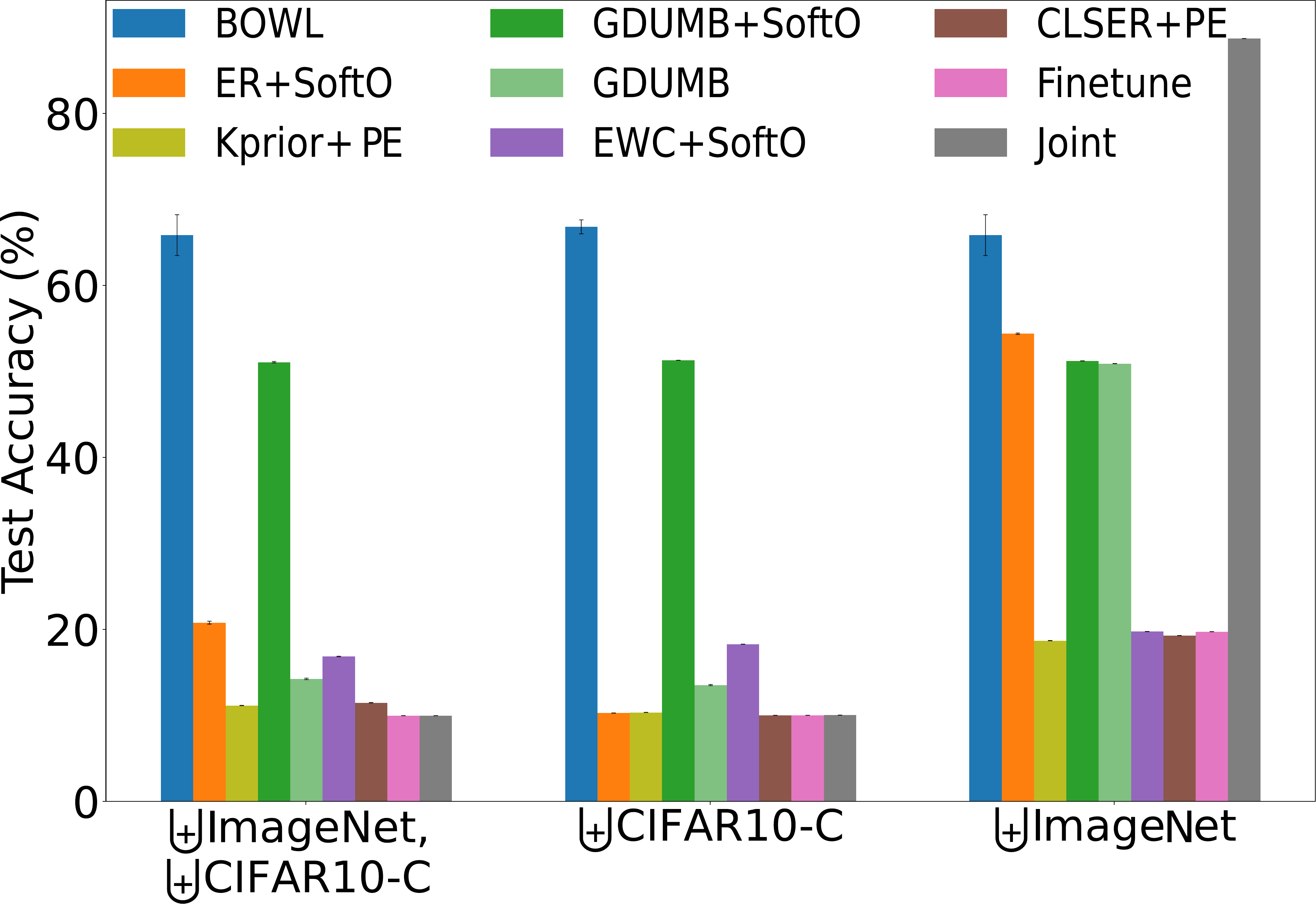}
    \captionof{figure}{\textbf{BOWL excels in open world learning scenarios in the presence of corrupted and unknown data.} Performance evaluation in terms of test accuracy across different setups when CIFAR-10 is combined with ImageNet and CIFAR10-C. BOWL demonstrates consistent performance across all the scenarios
    due to its ability to filter relevant data and
    informative sample selection.}
    \label{fig:bowl_open_world_learning}
    \end{minipage}
\end{figure*}

We conduct an ablation study on Split CIFAR-10 using the ResNet-18 model to analyze 
the utility of each design element, as introduced
in Section \ref{sec:bowl_modules}. 
Table \ref{tab:ablation_study_1} presents the results highlighting the importance of each module.

In the first row \textbf{i}, we show BOWL without the OoD module and retain the AL and CL components. 
BOWL is then trained on only 10581 samples - a significant reduction ($\approx$ 80\%) from CIFAR-10's dataset. 
This is attributed to the sample selection strategy in the AL module. 
The performance parity suggests two key insights: i) standardized benchmark datasets like CIFAR-10 contain significant redundancy, which can be overcome through sample selection, ii) the AL module effectively identifies the most informative samples, enabling efficient learning from a minimal dataset. 
The functionality of the OoD module is much more prominent for 
handling unknown future scenarios despite the minor trade-off,
especially in real-world scenarios where out-of-distribution data is common.
This will be strongly substantiated in \S \ref{sec:open_world_learning}. 
In row \textbf{ii}, we disconnect the Active Learning (AL) module from BOWL and use \emph{random sampling} of fixed-size queries to populate the memory buffer at the beginning of each task.
The test accuracy drops by $57\%$ in the first task itself and deteriorates further along the timeline.
This decline can be attributed to the homogeneity in the memory buffer on which the model is being trained.
When the Continual Learning (CL) module is disabled, as seen in row \textbf{iii}, BOWL clearly features the lowest final accuracy. 
This mainly occurs because of the model's tendency to catastrophically forget despite learning only on clean and important data-points. 
Without the CL module, there’s no mechanism to retain or consolidate past knowledge. As the model learns new tasks, the weights adapt primarily to the current task, overwriting representations for older tasks.
Even though the OoD module ensures that the model trains only on relevant data, and AL prioritizes informative data points, these alone cannot prevent the loss of past knowledge without a dedicated CL strategy.
The row \textbf{iv} exemplifies the performance of
the complete BOWL framework with the 3 modules by outperforming the other ablated variants.
The combination of OoD and AL ensures that the model learns from a smaller, high-quality subset of data without compromising accuracy. 
This supports the observation that a significant portion of CIFAR-10's data is redundant.
The CL module plays a critical role in retaining knowledge from past tasks. 
Merely optimizing for data quality does not address the challenge of address the challenge of retaining knowledge across tasks. 
While there is a natural decline in other cases, the drop is much less severe compared to case \textbf{iii}, where CL is absent. 
This demonstrates the importance of the CL module in mitigating catastrophic forgetting.

\subsection{Open World Learning}
\label{sec:open_world_learning}

The key strength of BOWL lies in 
extending any neural network with batch normalization layers to open world learning. 
To evaluate the efficiency of BOWL 
in such an open world, where 
the model encounters outliers and extraneous data points,
we corrupt CIFAR-10 with CIFAR-10-C \citep{hendrycks2019robustness} and include ImageNet \citep{imagenet}.
CIFAR-10-C contains systematically corrupted versions of CIFAR-10 images with 15 different types of corruptions (like noise, blur, weather effects). 
ImageNet has more complex scenes, backgrounds and some overlap in the features, but further contains non-relevant classes for CIFAR-10 classification.
Hence for a model to ensure high learning quality, it should discard samples from 
both the datasets.
We compare the average test accuracy
with other CL methods such as: 
ER, Kprior, GDUMB, EWC and CLSER. 
For a fairer comparison under open world assumptions, we further equip these CL methods
with two baseline OoD detection methods: SoftO \citep{softo} and Predictive Entropy (PE; \citet{predictive_entropy}). 
We have also included the \emph{joint} and \emph{finetune} training baselines, in order 
to get better insight into open world learning when trained naively.
We use a Resnet-18 \citep{resnet_2016} model to train on the datasets and employ a memory buffer of size 5000 for all methods that make use of replay. 

Figure \ref{fig:bowl_open_world_learning} illustrates the performance of BOWL in comparison with other standard CL algorithms.
ER, Kprior, EWC, CLSER and Finetune demonstrate marked degradation in their performance
when mixing data with different distributions, 
even when they augmented with respective OoD detection methods.
It is interesting to note that the combination of
GDUMB+SoftO outperforms ER+SoftO substantially, when the clean CIFAR-10 data is with mixed with CIFAR-10-C and ImageNet. 
Despite employing the same OoD measure, GDUMB's strategy to train from scratch makes the model less prone to corruption, using the resets to rectify the initial mistakes that ER cannot circumvent.
However, while GDUMB+SoftO improves the performance of GDUMB, it represents an ad-hoc combination that lacks grounding and falls significantly behind BOWL. When used without the SoftO component, GDUMB alone performs poorly, as can be seen in Figure \ref{fig:bowl_open_world_learning} (light green bar). This highlights GDUMB's fundamental inability to handle corruptions in the data without external assistance. In contrast, BOWL achieves superior performance through a principled, integrated approach that leverages batch normalization statistics rather than requiring separate OoD detection mechanisms.
BOWL displays observable reliability in its performance,
yielding $\approx60\%$ accuracy across different setups. \\
In the open world scenario, we can finally fully appreciate the importance of BOWL's three components:
i) BOWL's OoD mechanism
identifies and filters out both the corrupted CIFAR-10 and unrelated ImageNet samples, maintaining a focused learning objective on the core CIFAR-10 distribution.
ii) BOWL's active learning module ensures that the model trains only on informative
CIFAR-10 data by periodically querying from the relevant task pool.
iii) BOWL's continual learning strategy retains knowledge on previously learned samples.
Despite moving to an open world scenario, BOWL's final test accuracy aligns with the clean CIFAR-10 benchmark performance seen in Table \ref{tab:bowl_dataset_stats}.
The performance across different scenarios is consistent, especially compared to other methods that show high volatility in results. 

\section{Conclusion}
We have introduced a simple, yet effective, monolithic baseline for open world learning: BOWL. Our method leverages the statistics from the batch normalization layer to unify the three objectives of OoD detection, Active Learning, and Continual Learning synergistically. It detects outliers to discard irrelevant samples, achieves high learning speed by prioritizing data using informative queries, and mitigates forgetting by maintaining a dynamic memory buffer. 
The simplicity of BOWL arises from its reliance on diagonal covariance matrices from the batch normalization layer,
which is both a feature and simultaneous limitation, see Appendix \ref{app:limitations}.
Adopting BOWL can thus augment any standard neural network with batch normalization layers to make it a competent open world learner.
We anticipate future works to adopt advanced implementations of the present batch norm's simplified Gaussian and broad application to other tasks, such as regression or clustering. Overall, we envision BOWL to aid in meaningful assessments of open world learning. 

\section*{Ackowledgements}
This work supported by the Hessian Ministry of Higher Education, Research, Science and the Arts (HMWK; projects “The Third Wave of AI” and “The Adaptive Mind”). It further benefited from the Hessian research priority programme LOEWE within the project “WhiteBox”.

\bibliography{arxiv/collas2025_conference}
\bibliographystyle{arxiv/collas2025_conference}

\onecolumn

\appendix
\section{Appendix}
\label{app:appendix}

The appendix complements the main body with additional details on:
\begin{enumerate}
    \item[A.1] The required elements and a loose definition of the term open world learning
    \item[A.2] A derivation of BOWL's out of distribution score
    \item[A.3] Data sequences, training hyper-parameters 
    evaluation measures and computational resources used in experiments
    \item[A.4] A discussion on limitations and prospects
\end{enumerate}

\subsection{The Elements of Open World Learning}
\label{app:OWLL_definition}

In this section, we provide a more detailed intuition behind the components of open world learning. Using natural language to ease understanding, \cite{Bendale_2016_CVPR} define open world recognition (learning) as:
\begin{quote}
    ``In open world recognition the system must be able to recognize objects and associate them with known classes while also being able to label classes as unknown. These “novel unknowns” must then be collected and labeled (e.g. by humans). When there are sufficient labeled unknowns for new class learning, the system must incrementally learn and extend the multi-class classifier, thereby making each new class “known” to the system. Open World recognition moves beyond just being robust to unknown classes and toward a scaleable system that is adapting itself and learning in an open world.'' - \emph{Bendale and Boult, Towards Open World Recognition}
\end{quote}

Without quoting the entire formal statement, (please refer to \cite{Bendale_2016_CVPR} Definition 1), the mathematical description can intuitively be summarized as:
\begin{enumerate}
    \item An \emph{open set recognition function} that involves a novelty detector to determine whether any result from the recognition function is from an unknown class.
    \item An \emph{active labelling function}, typically a human oracle in supervised learning, to label any unknown data points and, if necessary, extend the set of existing known classes.
    \item An \emph{incremental learning function} that trains the system by adding the new data points and respectively continuing to train the recognition function.
\end{enumerate}

BOWL follows the general concept of open world learning and provides the first monolithic baseline to empower experimental progress and transparent comparison. As such, it further disambiguates above definition in places where the exact implementation allows for a vague interpretation. 
In favor of generality, BOWL thus encapsulates a broader interpretation of steps two and three, in the spirit of the desiderata of lifelong learning \citep{mmundt-2020:owll}.

Specifically, in the active labeling function of step 2, BOWL includes an actual \emph{active learning query}. In other words, rather than simply labeling all data that the out of distribution detector has detected as unknown, the active query step further gauges \emph{informativeness} and \emph{relatedness of the data}. This is important, because the open world learner should leverage examples that are remotely related to what the objectives are, rather than introducing any sorts of novel points. If we think of the novelty detector giving a high score to arbitrary unseen noise, it is important to gauge whether inclusion of this example is expected to reduce prospective loss. In turn, such inclusion of an active learning query further significantly reduces the amount of (redundant) data used for continuous training of open world learning system.

In addition, and most importantly, we place the strict requirement on the incremental learning step to avoid concatenation of data. In fact, existing algorithms that tackle some form of open world learning \citep{Joseph_2021_CVPR} place a large emphasis on the novelty detector and active data inclusion, yet they continue extending the dataset in the spirit of traditional active learning. In the spirit of continual learning and overall real-world system plausibility, we avoid retention of all old data and thus need to combat any expected catastrophic forgetting \citep{MCCLOSKEY1989109}.

In the easiest sense, our three above points thus remain the same, but are augmented with the specific challenge of gauging data informativeness and mitigating common catastrophic interference. 

\subsection{Derivation of Out of Distribution Score}
\label{app:ood_derivation}
In this appendix we show how decision rules for outlier detection based on $\eta_0$ and $\eta_1$
The log probability density function of the multivariate t-distribution with
mean $\mathbf{\mu}$,
variance $\frac{\nu}{\nu - 2} \mathbf{\Sigma}$,
degrees of freedom $\nu$,
and dimensionality $d$
is as follows:
\begin{equation}
    \ln P(\mathbf{x}) = 
    -\ln \Gamma \left( \frac{\nu + d}{2} \right)
    -\ln \Gamma \left( \frac{\nu}{2} \right)
    -\frac{1}{2}\ln \det(\nu \pi \mathbf{\Sigma})
    - \frac{\nu + d}{2} \ln \left(
    1 + \frac{1}{\nu} (\mathbf{x} - \mathbf{\mu})^T \mathbf{\Sigma}^{-1} (\mathbf{x} - \mathbf{\mu})
    \right)
    \label{eqn:t-dist}
\end{equation}
As a belief about future observations of $\mathbf{x}$ this corresponds to assuming that $\mathbf{x}$ is distributed according to a multivariate normal distribution, the mean and covariance of which were estimated from $\nu$ observations of previous $\mathbf{x}$s.
In the limit $\nu \to \infty$ we recover the normal distribution
\begin{equation}
    \ln P(\mathbf{x}) = 
    -\frac{1}{2} (\mathbf{x} - \mathbf{\mu})^T \mathbf{\Sigma}^{-1} (\mathbf{x} - \mathbf{\mu}) + C
    \label{eqn:normal}
\end{equation}
where $C$ is a normalizing constant and does not depend on $\mathbf{x}$.

If we perform a Bayesian hypothesis comparison between the inlier hypothesis ``the intermediate activations $\mathbf{x}$ are drawn from a normal distribution with mean $\mu$ and covariance $\mathbf{\Sigma}$" and the outlier hypothesis ``the intermediate activations are drawn from a uniform distribution",
we obtain a posterior log odds ratio of
\begin{equation}
    \eta_0 = \frac{1}{2} (\mathbf{x} - \mathbf{\mu})^T \mathbf{\Sigma}^{-1} (\mathbf{x} - \mathbf{\mu}) + C_0
\end{equation}
in favour of the outlier hypothesis, where $C_0$ includes both a normalizing constant and the prior log odds in favour of the outlier hypothesis.

The observed failure mode of using large values of $\eta_0$ in a decision rule for outlier detection is that outlying data which produces anomalously small intermediate activations is accepted as inlying.
One way of interpreting this is to note that our inlier hypothesis corresponds to very high confidence that the covariance of activations is $\mathbf{\Sigma}$,
(the limit $\nu \to \infty$ corresponding to having estimated $\mathbf{\Sigma}$ from arbitrarily many prior samples of $\mathbf{x}$).
If we now ask what our outlier hypothesis corresponds to, we see that it corresponds to very high confidence that outliers produce activations with a very large covariance. (That is, the limit of equation \ref{eqn:normal} as $\mathbf{\Sigma}$ becomes large is the uniform distribution).
From this point of view it is reasonable that our decision rule malfunctions for outliers which have anomalously small intermediate activations:
we have implicitly assumed that such outliers do not exist.

Intuitively we want our outlier hypothesis to instead say something like "the intermediate activations are drawn from a normal distribution with a covariance about which we are highly uncertain".
We can achieve this by taking the opposite limit of equation \ref{eqn:t-dist} to that which constructed equation \ref{eqn:normal}, $\nu \to 0$.
Doing so produces an unnormalized (and indeed, like the uniform distribution, unnormalizable) log probability density of
\begin{equation}
    \ln P(\mathbf{x}) = 
    - \frac{d}{2} \ln \left( (\mathbf{x} - \mathbf{\mu})^T \mathbf{\Sigma}^{-1} (\mathbf{x} - \mathbf{\mu}) \right)
    \label{eqn:uninformative}
\end{equation}
If we now perform Bayesian hypothesis comparison with equation \ref{eqn:normal} as the inlier hypothesis and \ref{eqn:uninformative} as the outlier hypothesis,
we obtain a posterior log odds ratio of
\begin{equation}
    \eta_1 = \frac{1}{2} (\mathbf{x} - \mathbf{\mu})^T \mathbf{\Sigma}^{-1} (\mathbf{x} - \mathbf{\mu})
    - \frac{d}{2} \ln \left( (\mathbf{x} - \mathbf{\mu})^T \mathbf{\Sigma}^{-1} (\mathbf{x} - \mathbf{\mu}) \right)
    + C_1
    \label{eqn:eta_1}
\end{equation}
in favour of the outlier hypothesis, where $C_1$ is again the combination of a normalizing constant and the prior log odds ratio in favour of the outlier hypothesis.
The reader will observe that, unlike $\eta_0$, $\eta_1$ is large for both large and small magnitude values of $(\mathbf{x} - \mathbf{\mu})$.

We set the threshold $\tau$ using the bootstrap method \citep{nalisnick2019detecting}. Although for the initial setting a held-out validation dataset is used, in our case we use the samples in the memory buffer. Similar to \cite{nalisnick2019detecting}, we sample $K$ ‘new’ data sets $X_{k=1}^{|\mathcal{M}|}$ of size M from the memory buffer $\mathcal{M}$ and then calculate $\eta_1k$ for the $kth$ bootstrap set. We then calculate $\alpha$-quantile over the $\eta_1k$ distribution to get the overall threshold estimate $\tau$.

\subsection{Dataset Sequences and Training Hyper-parameters} 
\label{app:training_configs}

\subsubsection{Split CIFAR-10}
We divide CIFAR-10 \citep{cifar10_Krizhevsky09} (consisting of 50k color images equally balanced across 10 classes of $32\times32$ resolution) into 5 disjoint timesteps (classes $\{2, 5\} \rightarrow [0, 6] \rightarrow [1, 7] \rightarrow [3, 8] \rightarrow [4, 9]$ ) at timestep $t=0 \rightarrow t=1 \rightarrow t=2 \rightarrow t=3 \rightarrow t=4$ i.e Split CIFAR-10. We use a ResNet-18 \citep{resnet_2016} to train on data with labels that come under $\{\ldots\}$ braces with two output nodes for classification at timestep $t=0$. We use SGD with a learning rate of $0.1$, a momentum value of $0.9$ and weight decay of $0.0005$ and train for $120$ epochs with mini-batch size $b=256$. This training is performed under traditional setup. 
For BOWL, the memory buffer is filled with random samples from the training data (availability of old training data is assumed only at this initial point in time). 
The model first discards irrelevant data at timestep $t=1$ with batch size $b=8$. 
To enable open world learning of the model to the new data available at incremental timestep $t>0$, we expand the number of output nodes of the classifier to the number of classes detected at timesteps $t$ by the number of unique class labels detected in the queryable pool. The expanded model is then used for continual learning on the upcoming disjoint tasks. 
The accepted data is available to actively query and learn from. For one loop of active query, acquired batch of size 256, supersedes unimportant samples in the memory buffer. 
The model is then trained in continual fashion using the same configuration as stated earlier but for $2$ epochs. 
The active query loop iterates until the queryable pool is vacant. We measure the test accuracy at the end of training for each timestep on classes seen thus far. We report results wit standard deviations over 5 runs.

\subsubsection{Split ImageNet-200}
We divide ImageNet-200 (a subset of ImageNet consisting of 200 classes) into 10 disjoint timesteps. The dataset contains high-resolution images selected from ImageNet through their WordNet IDs (e.g., n01443537, n01944390). The classes are divided by sequentially grouping 20 WordNet IDs, where each group represents a timestep, resulting in classes being distributed from $t=0$ to $t=9$. We use an EfficientNet-B2 \citep{tan2019efficientnet} to train on data with labels that come under the first timestep with 20 output nodes for classification at timestep $t=0$. We use SGD optimizer with a learning rate of $0.01$, a momentum value of $0.9$, and weight decay of $1\times10^{-4}$, coupled with a cosine annealing learning rate scheduler that decays from the initial learning rate to $1\times10^{-5}$ across the training duration. We train for $300$ epochs with mini-batch size $b=128$.

\subsubsection{Presence of OoD and corrupted data: CIFAR-10-C and ImageNet}
\textbf{CIFAR-10-C} \citep{hendrycks2019robustness}: We apply 3 corruptions namely: impulse noise, Gaussian noise, shot noise to all the classes of  CIFAR-10 \citep{cifar10_Krizhevsky09} dataset. Out of the 75 available noise types, we chose these 3 as they simulate common noise additions that occur more often that can affect model performance when trained on. Gaussian noise especially can be more harmful depending on their severity. We generate 50000 images for each noise type equally balanced across all classes. The image meta-information such as width, height and channel remain the same.
\\
\textbf{ImageNet} \citep{imagenet}: We use a subset of the ImageNet dataset with labels $0, 1, 2, 3, 4, 5, 6, 7, 8, 9$ and about 1000 images per class.  We discard remaining subset. These clean images from ImageNet act as the inputs which the model needs to discard in order focus on relevant ``in-distribution'' data. 

We then interleave the corrupted data(CIFAR-10-C) and the subset of ImageNet dataset with the original CIFAR-10 dataset to simulate data for open world learning.


\subsubsection{Computing the cosine similarity}
We employed a strategic approach to handling memory constraints when computing pairwise similarities across the datasets. 
The core challenge lies in managing the quadratic memory growth inherent in similarity matrix calculations, where processing $N$ samples requires an $N\times N$ matrix that often exceeds available GPU memory.
By keeping the normalized data resident in GPU memory throughout the computation while systematically transferring computed similarity chunks to CPU memory, the implementation achieves scalability without compromising performance. This approach is particularly effective because it leverages the GPU's computational power for the intensive matrix multiplication operations while utilizing the CPU's larger memory capacity for storage of intermediate results.

\subsubsection{Details on comparison of methods}
We benchmark against 
various methods. 
1) Experience Replay (ER) \citep{ER_2019, rolnick2019experience}: mitigates forgetting through a random memory buffer, yet follows the conventional strategy to interleave it with all available current task data - a useful baseline to assess the focus on forgetting,
2) Elastic Weight Consolidation (EWC) \citep{ewc}:
is a method that regularizes weights relevant to the previous task when training on the current task,
3) GDUMB \citep{gdumb}: although not particularly designed for open world learning, it is a good contender against a plethora of continual learning approaches. Like BOWL, GDUMB trains only on a memory buffer but populates it randomly. Hence, it serves as a suitable baseline for assessing our active queries' quality. 
4) CLSER \citep{arani_complementarylearning}:
is a dual memory experience replay (ER) method to mitigate catastrophic forgetting. The method leverages episodic memory and regularization to efficiently learn new tasks while retaining previously learned knowledge.
5) Kprior \citep{kpriors_khan2021}: stores examples relevant to the previous task and performs functional as well as weight regularization while training on the current tasks,
In addition to the above, we also compare it against simple ``Finetune" where the datasets are introduced sequentially to mimic an incoming data stream at different timesteps.
The training configuration including model backbone, opimizer, learning rate, batch size, epochs is fixed across all the methods and 
timesteps.

\subsubsection{Evaluation measures}
We primarily report test accuracy at the end of the last task to evaluate and compare the performance of BOWL, as sketched in the main body, in addition to reporting the overall number of data points used for training. 

\textbf{Number of observed data points (\#ODP)}:
For each task, we calculate the total number of observed data points by aggregating unique samples queried during periodic buffer updates. At each update step, BOWL queries a batch of samples from the pool.
We then integrate these queried samples with 
the old samples in the buffer and select top scoring samples
of the size of memory buffer.
We then note the number of ``new'' samples that is 
populated in the buffer for training the model.
We consider such samples only once in our final sample count to provide a fair comparison with other methods.

\textbf{Average Classification Accuracy}: 
We measure the performance of a model on the cumulative test dataset after learning on the train data at each incremental timestep.
At each timestep $t$, after training the model on train dataset $D_{t}^{train}$, we calculate the accuracy $a_t$ of the model on the test dataset $D_{t}^{test}$.
The final average accuracy after $T$ timesteps is given as: 
\begin{equation}
    A_T = \frac{1}{T}\sum_t^{T} a_t    
\end{equation}
In context of BOWL, average classification accuracy provides a good estimate of the overall system's balance between encoding new information while maintaining prior learned knowledge.


\subsection{Limitations and Prospects}
\label{app:limitations}
We provide additional discussions on the limitations and future prospects for BOWL.
BN layers normalize the pre-activations of a layer using the statistics from mini-batches of data to yield zero mean, unit variance and diagonal covariance. 
We emphasize that the diagonal covariance assumption is a simplification (pertaining to de-correlations), yet BOWL delivers a strong simple monolithic baseline using just the diagonal values. We list other limitations as follows:\\

\textbf{Memory buffer size}: As in all memory-based continual learning methods, BOWL requires a good estimate on the size of the memory buffer to store data. At present, all approaches, including BOWL, thus motivate buffer size from a storage and compute constraint and make a choice a priori. The memory requirement can however increase or vary with time and one can investigate methods to dynamically allocate memory, depending on both how informative novel data is and the expected performance of the model. The a priori choice is thus a current limitation, requiring some intuition of task complexity, and the dynamic extension of memory size an enticing future prospect. \\

\textbf{Generalization and application beyond supervised open world learning}: Already in the current form, BOWL's OoD module detects the outlier on small batch size in an unsupervised fashion and without any prior training on out-of-distribution data, using only population statistics maintained in the batch-norm layers. Similarly, the active query module acquires novel samples using entropy obtained from the BN layer activations weighted with data similarity. This again is done without any information about the label or any form of supervision. In essence, one can thus trivially extend BOWL to unsupervised learning, reinforcement learning, or other prediction tasks, such as e.g. semantic segmentation. We thus expect BOWL to fuel further development of open world learning beyond the supervised examples in this paper.

\end{document}